\documentclass[letterpaper]{article} % DO NOT CHANGE THIS
\usepackage{aaai25}  % DO NOT CHANGE THIS
\usepackage{times}  % DO NOT CHANGE THIS
\usepackage{helvet}  % DO NOT CHANGE THIS
\usepackage{courier}  % DO NOT CHANGE THIS
\usepackage[hyphens]{url}  % DO NOT CHANGE THIS
\usepackage{graphicx} % DO NOT CHANGE THIS
\usepackage{natbib}  % DO NOT CHANGE THIS AND DO NOT ADD ANY OPTIONS TO IT
\usepackage{caption} % DO NOT CHANGE THIS AND DO NOT ADD ANY OPTIONS TO IT
\usepackage{amsmath} 
\usepackage{booktabs}
\usepackage{comment}
\usepackage{algorithm}
\usepackage{algorithmic}

\usepackage{newfloat}
\usepackage{listings}
\DeclareCaptionStyle{ruled}{labelfont=normalfont,labelsep=colon,strut=off} % DO NOT CHANGE THIS
\floatstyle{ruled}
\newfloat{listing}{tb}{lst}{}
\floatname{listing}{Listing}
\title{Weighted Poisson-disk Resampling on Large-Scale Point Clouds}

\def\correspond{%
    \ifnum\value{eqfn}=1%
    \footnote{Corresponding authors.}%
    \setcounter{eqfn}{\value{footnote}}%
    \else%
    \footnotemark[\value{eqfn}]%
  \fi%
}%

\author{
    Xianhe Jiao\textsuperscript{\rm 1}\equalcontrib, 
    Chenlei Lv\textsuperscript{\rm 2}\equalcontrib, 
    Junli Zhao\textsuperscript{\rm 1}\correspond, 
    Ran Yi\textsuperscript{\rm 3},\\ 
    Yu-Hui Wen\textsuperscript{\rm 4}, 
    Zhenkuan Pan\textsuperscript{\rm 1}, 
    Zhongke Wu\textsuperscript{\rm 5}, 
    Yong-jin Liu\textsuperscript{\rm 6}\correspond
}

\affiliations{
    \textsuperscript{\rm 1}College of Computer Science and Technology, Qingdao University;
    \textsuperscript{\rm 2}College of Computer Science and Software Engineering, Shenzhen University;
    \textsuperscript{\rm 3}Department of Computer Science and Engineering, Shanghai Jiao Tong University;
    \textsuperscript{\rm 4}School of Computer Science \& Technology, Beijing Jiaotong University;
    \textsuperscript{\rm 5}School of Artificial Intelligence, Beijing Normal University;
    \textsuperscript{\rm 6}Department of Computer Science and Technology, MOE-Key Laboratory of Pervasive Computing, Tsinghua University\\   
    \{zjl@qdu.edu.cn, liuyongjin@tsinghua.edu.cn\}
}

\usepackage{bibentry}

\begin{document}

\maketitle

\begin{abstract}
For large-scale point cloud processing, resampling takes the important role of controlling the point number and density while keeping the geometric consistency. % in related tasks. 
However, current methods cannot balance such different requirements. Particularly with large-scale point clouds, classical methods often struggle with decreased efficiency and accuracy. To address such issues, we propose a weighted Poisson-disk (WPD) resampling method to improve the usability and efficiency for the processing. We first design an initial Poisson resampling with a voxel-based estimation strategy. It is able to estimate a more accurate radius of the Poisson-disk while maintaining high efficiency. Then, we design a weighted tangent smoothing step to further optimize the Voronoi diagram for each point. At the same time, sharp features are detected and kept in the optimized results with isotropic property. Finally, we achieve a resampling copy from the original point cloud with the specified point number, uniform density, and high-quality geometric consistency. Experiments show that our method significantly improves the performance of large-scale point cloud resampling for different applications, and provides a highly practical solution.
\end{abstract}

\section{Introduction}

{W}{i}th the development of 3D scanning technology, 3D point clouds are widely collected and gradually {become} %the most 
a popular data representation in 3D vision tasks. Compared to 2D images, 3D point clouds possess comprehensive geometric information, enabling precise spatial data analysis. Although point clouds possess many {outstanding} properties, there are still some limitations {that} constrain their usage. 
Specifically, point clouds scanned from large-scale scenes often entail substantial data volumes, which reduce computational efficiency in related applications. In addition, the density of a raw point cloud is typically non-uniform, %and the drawback 
{which} significantly impairs the performance of certain downstream applications that are highly sensitive to the quality of point distribution. Therefore, a resampling step is necessary.

\begin{figure}[t]
\centering
\includegraphics[width=0.9\columnwidth]{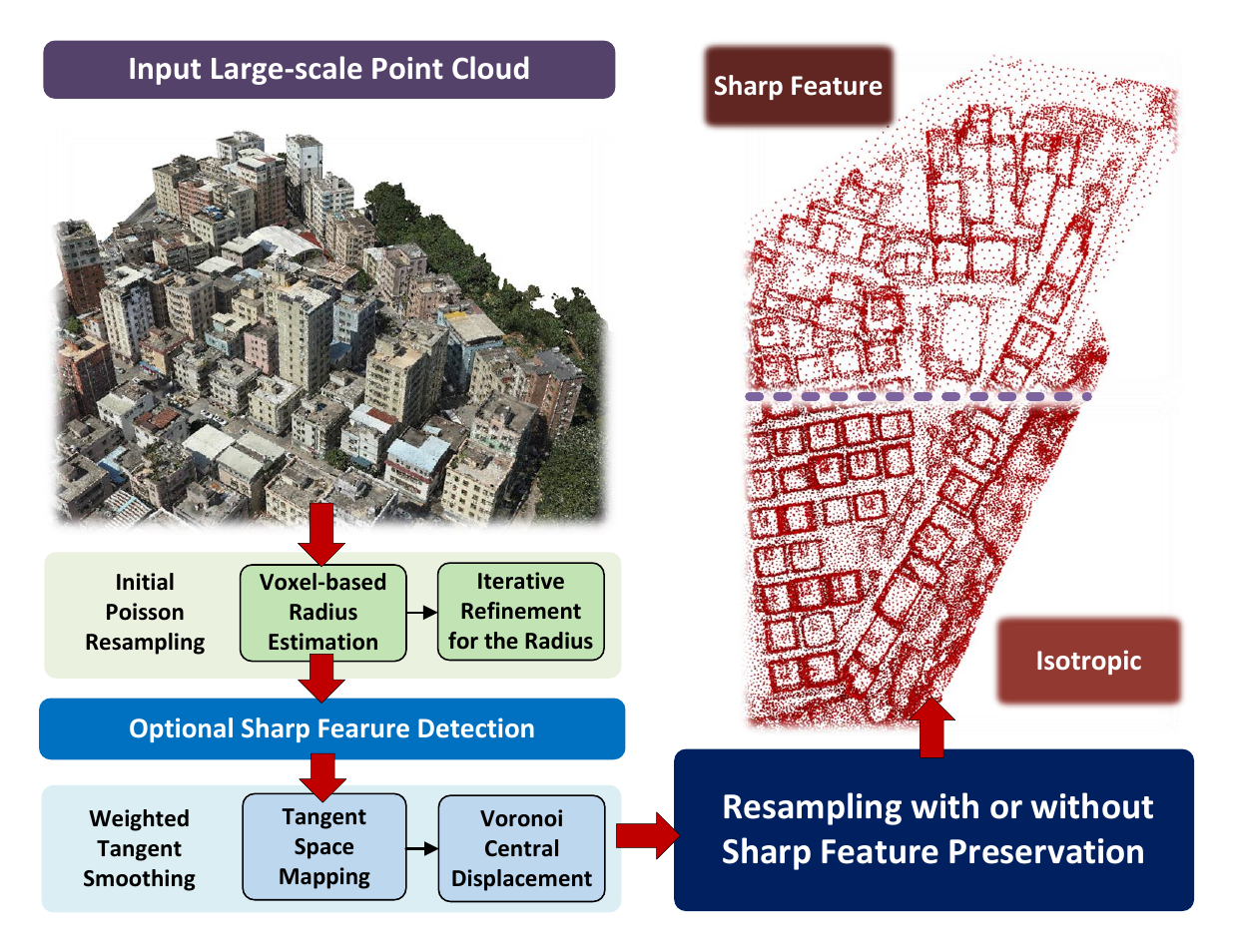} 
\caption{The pipeline of the proposed WPD on UrbanBIS~\cite{Yang2023UrbanBIS}.}
\label{fig:pipeline}
\end{figure}

The resampling step aims to control the number of points while optimizing their distribution. Currently, the mainstream solutions include Poisson-disk resampling~\cite{corsini2012efficient} and variants of the farthest point sampling (FPS)~\cite{schlomer2011farthest, lv2021approximate}. The former has been selected as a standard function of the well-known software MeshLab~\cite{cignoni2008meshlab}. It can effectively improve the distribution of point clouds while reducing the data volume. The drawback is that the radius estimation of {the} Poisson-disk is not accurate. It cannot precisely control the number of output points. For the variants of FPS, the advantage lies in their ability to precisely control the number of points with simple implementation. Such methods are widely used in point cloud-based deep learning methods~\cite{qi2017pointnet++, zhao2021point}. However, they have lower efficiency for large-scale point clouds, and their optimization capability for local neighborhoods is also limited.

To address the above{-}mentioned issues, we propose a weighted Poisson-disk (WPD) resampling method for large-scale point cloud processing. It contains two parts: initial Poisson resampling and weighted tangent smoothing. The initial Poisson resampling is an improved version based on the {traditional} Poisson-disk resampling. Based on a voxel-based analysis, it estimates a more accurate Poisson-disk radius according to the specified number of points. A weighted tangent smoothing is used to further control the point number while optimizing the local neighborhoods. It efficiently establishes the isotropic distribution for point clouds. Additionally, we provide an optional function for {preserving sharp features}. With a sharp feature detection~\cite{jiao2023msl}, WPD resampling effectively balances varying requirements across downstream applications, including visualization and semantic feature learning. In Figure~\ref{fig:pipeline}, we show the pipeline of the proposed WPD resampling method. {The contribution of this paper is three folds}:

\begin{itemize}

    \item We design an initial Poisson resampling step that is an improved version of {the} {traditional}  Poisson-disk resampling. It utilizes a voxel-based estimation to improve the accuracy of Poisson-disk radius according to the specified point number.
    
    \item We present a weighted tangent smoothing to further optimize point distributions. Combined with the initial Poisson resampling step, it can implement isotropic resampling for large-scale point clouds with high efficiency and better geometric consistency.
    
    \item We provide an optional function for sharp feature {preservation} during the resampling. It combines robust sharp feature analysis with different sampling {rate} settings, to ensure compatibility %to be compatible 
    with WPD resampling. 
    
\end{itemize}

\section{Related Work}\label{sec:rw}

For point cloud resampling, related solutions can be summarized into three categories: local uniform optimization, geometric feature-based and semantic feature-driven methods.

\textbf{Local uniform optimization} attempts to improve the quality of point distribution based on local distances. The representative methods include farthest point sampling (FPS)~\cite{moenning2003fast, schlomer2011farthest, lv2021approximate}, consolidation~\cite{lipman2007parameterization, Huang2009}, Poisson-disk resampling~\cite{corsini2012efficient}, Voronoi diagram optimization~\cite{liu2009centroidal, Chen2018La}, and Laplace graph~\cite{Luo2018, Chen2018, Qi2019, Zeng2020}. FPS-based {schemes have} been widely used for point cloud pre-processing. The various modified versions of FPS~\cite{schlomer2011farthest, lv2021approximate} achieved improvements in both geometric consistency preservation and feature enhancement. For point cloud-based consolidation, Huang \emph{et~al.}~\cite{Huang2009} proposed a weighted locally optimal projection (WLOP) operator to optimize local point distributions. Poisson-disk resampling~\cite{corsini2012efficient} is a practical resampling solution {that} has been 
integrated into MeshLab as a standard function. The proposed WPD resampling is inspired by the solution.

\textbf{Geometric feature-based methods} implement point cloud resampling while balancing geometric feature preservation. Some classical solutions include DSO (Discrete Shape Operator) feature-based simplification~\cite{Lee2011}, normal vector-driven simplification~\cite{shi2011adaptive}, sharp feature keeping \cite{Huang2013, Benhabiles2013}, curvature adaption \cite{Liu2013, lv2022adaptively}, and saliency-based resampling \cite{Ding2019}. They consider geometric feature keeping in resampling{,} which can improve the quality of the geometric consistency. For manifold distribution property keeping, some methods design the intrinsic resampling schemes based on geodesic distance, including intrinsic resampling~\cite{lv2022intrinsic} and geodesic Voronoi diagrams\cite{wang2015intrinsic, liu2017constructing}. Such {a method} can output {high-quality} resampling result{s with} better consistency to the original 3D surface. However, the drawback of {these} methods is the excessively huge time cost for feature analysis, which limits their application for large-scale point clouds.

\textbf{Semantic feature-driven methods} consider the semantic analysis during resampling. {Their} greatest advantage lies in the ability to reconstruct missing local geometric information from the raw point cloud. The representative solutions include FoldingNet~\cite{yang2018foldingnet}, KCNet~\cite{shen2018mining}, PU-Net~\cite{yu2018pu}, SampleNet~\cite{dovrat2019learning}\cite{lang2020samplenet}, PAT~\cite{yang2019modeling}, CPL~\cite{nezhadarya2020adaptive}, MOPS-Net~\cite{qian2020mops}, PIE-NET~\cite{wang2020pie}, PointASNL~\cite{yan2020pointasnl}, SK-Net~\cite{wu2020sk}, etc. %, 
These methods resample the points with sensitive characteristics to facilitate effective semantic-based analysis. However, the performance of these solutions depends heavily on the sample distribution of the training dataset, which can lead to potential instability. Furthermore, semantic-driven sampling often fails to achieve a consistently reliable uniform distribution. In practice, similar approaches still require FPS strategy to improve the density. {Benefiting} from the sharp feature keeping, {our} WPD resampling method can enhance the geometric details that correspond to the semantic features. In {the} following parts, we introduce the implementation details.

\section{Methodology}\label{Methodology}

\textbf{Overview.} 
In this paper, we aim to develop a resampling method that resamples a large-scale input point cloud into a new one with {a} specified number of points and optimized point distribution. {Our} proposed WPD resampling method {consists of two stages:} the initial Poisson resampling and weighted tangent smoothing. The initial Poisson resampling employs a voxel-based estimation to improve the accuracy of point control while keeping the uniform distribution and sampling efficiency. The weighted tangent smoothing further enhances the point distribution to approach isotropic requirement and output resampling result{s} with {an} accurate number of points. As an option, we offer a sharp feature-sensitive dynamic resampling scheme that enhances the proportion of edge points in the resampling results with dynamic densities, thus reinforcing sharp features. In {the} following parts, we illustrate the implementation details.

\subsection{Initial Poisson Resampling}

\textbf{Preliminaries on Poisson-disk Resampling.} The {traditional} Poisson-disk resampling~\cite{corsini2012efficient} implemented by MeshLab is based on an intuitive assumption that relates surface area to the number of points. Based on the accumulation of the local area corresponding to each point, we can estimate the radius of each point that controls the number of points while maintaining {uniformity}. Let $P$ {represents} the input point cloud, {and} $n$ {represents} the specified number of resampling points. {Then}{,} the mentioned assumption can be formulated as:
\begin{equation}
S_P = \lambda n\pi r^2,
\label{e1}
\end{equation}
where $S_P$ represents the surface area of $P$, {and} $r$ is the radius of a local region. We accumulate a local region consisting of $n$ points, such that the sum of these points equals {to} $S$. Since each small region overlaps with others, a decay factor $\lambda$ is introduced. By reversing this assumption, if we know the surface area corresponding to the point cloud, then by controlling the sampling radius, we can accurately control the resampling point number while keeping the uniform distribution.

The original Poisson-disk resampling estimates the $S_P$ by the area of bounding box{, which} is represented as:
\begin{equation}
S_P \approx lh+wh+lw,
\label{e2}
\end{equation}
where $l,h,w$ are the length, width, and height of the bounding box, and $S_P$ is estimated as half {the} surface area of the bounding box. Clearly, it is a rough estimation. Especially when the point cloud exhibits significant curvature changes, the resampling cannot output result{s} with {an} accurate point number. To {address this} issue, we propose a voxel-based radius estimation to improve the accuracy.

\begin{figure}[t]
\centering
\includegraphics[width=0.9\columnwidth]{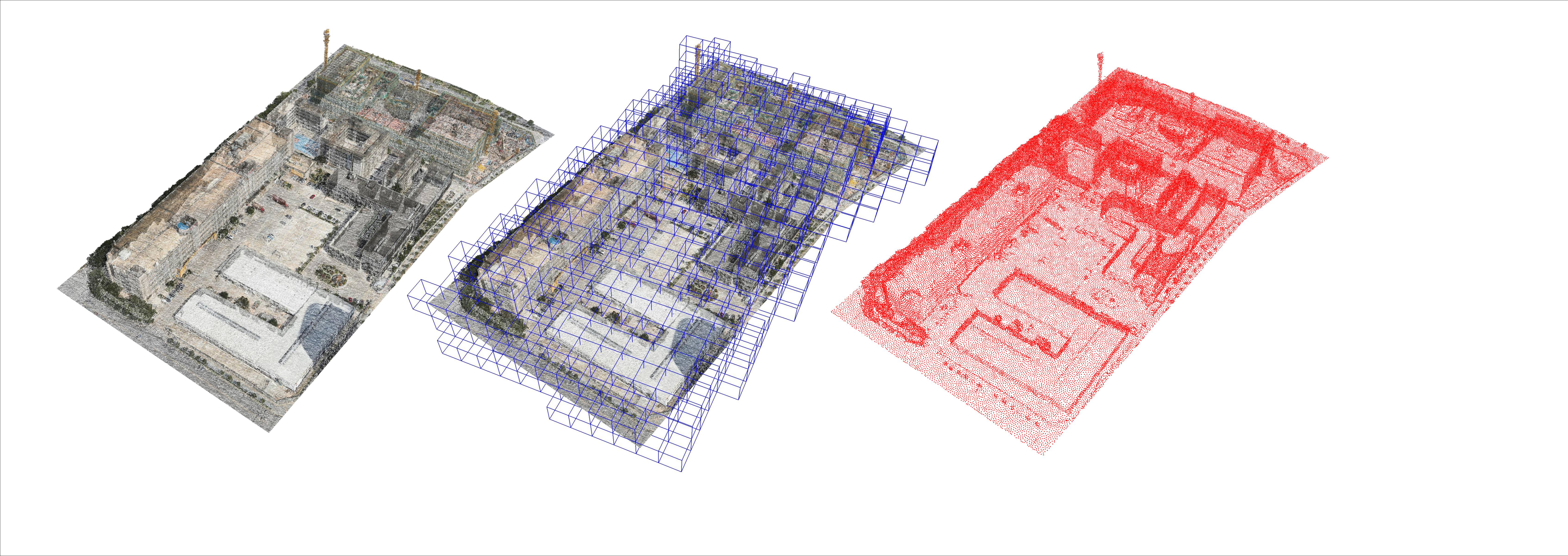}
\caption{Voxel-based radius estimation and its corresponding initial resampling result for a large-scale scene. Left: original point cloud; middle: voxelization; right: initial Poisson resampling result.}
\label{泊松盘半径对比}
\end{figure}

\textbf{Voxel-based Radius Estimation.} 
{The key to enhancing} the accuracy of the radius estimation is providing an efficient estimation of the surface area of {the} original point cloud. The most intuitive solution is to reconstruct a mesh from the point cloud and then calculate {its} surface area. However, this significantly increases {the} computational overhead. We introduce a more efficient solution that estimates surface area based on accumulated voxel-based surfaces. 

{Specifically,} we {first} implement the voxelization for input point cloud, the voxel length $l_v$ is set to $0.05 \times max\{l,h.w\}$ by default. Then, we estimate $S_P$ as
\begin{equation}
S_P \approx ml_v^2,
\label{e3}
\end{equation}
where $m$ is the voxel number, $ml_v^2$ means that we accumulate each face area from related voxel to represent the $S_P$. Since most voxels have only one face corresponding to the region of the point cloud. Based on the Eq.~\eqref{e1} and Eq.~\eqref{e3}, we achieve the new radius computed by
\begin{equation}
r = l_v\sqrt{m/(\lambda n\pi)},
\label{e4}
\end{equation}
where $\lambda$ is set to 0.68 according to the experimental results. The voxel-based radius estimation more effectively {considers} the morphology of the point cloud, thereby enhancing the accuracy of surface area computation. Consequently, the accuracy of radius estimation is improved, which indirectly improves the ability to control the number of points. Once the radius is defined, a more accurate Poisson-disk resampling can be implemented.

\textbf{Iterative Refinement.} {Although} the voxel-based radius estimation improves the accuracy {of} the resampling, there is still a discrepancy between the number of output points and the specified number. To facilitate subsequent processing, we {design} an iterative refinement step to further optimize the number of resampling points. We iteratively adjust the radius based on the error between the {number of} sampling points and the specified number, ultimately producing a more accurate result, {i.e.,} the resampling error is controlled under $5\%$. The implementation of iterative refinement can be formulated as:
\begin{equation}
R_n=1-|P'|/n,
\label{e5}
\end{equation}
\begin{equation}
r'=\left\{\begin{array}{l}r(1-\frac1{\theta_1}R_n),\;R_n>0\\r(1-\frac1{\theta_2+\mu}R_n),\;R_n<-0.05\end{array}\right.
\label{e6}
\end{equation}
where $|P'|$ represents the number of resampling points{.} 
We iteratively refine the radius $r$ by scaling parameters $\theta_1,\theta_2$, and $\mu$ ($\theta_1=1.8,\theta_2=3.0, \mu=(|P'|-n)/2$). By fine-tuning the sampling radius, we can further optimize the number of sampling points while maintaining the uniform point distribution. Since the process does not involve complex tangent space mapping or distance optimization, it only performs the original Poisson-disk sampling in each iteration, making the method highly efficient.

\begin{figure}[t]
\centering
\includegraphics[width=0.95\columnwidth]{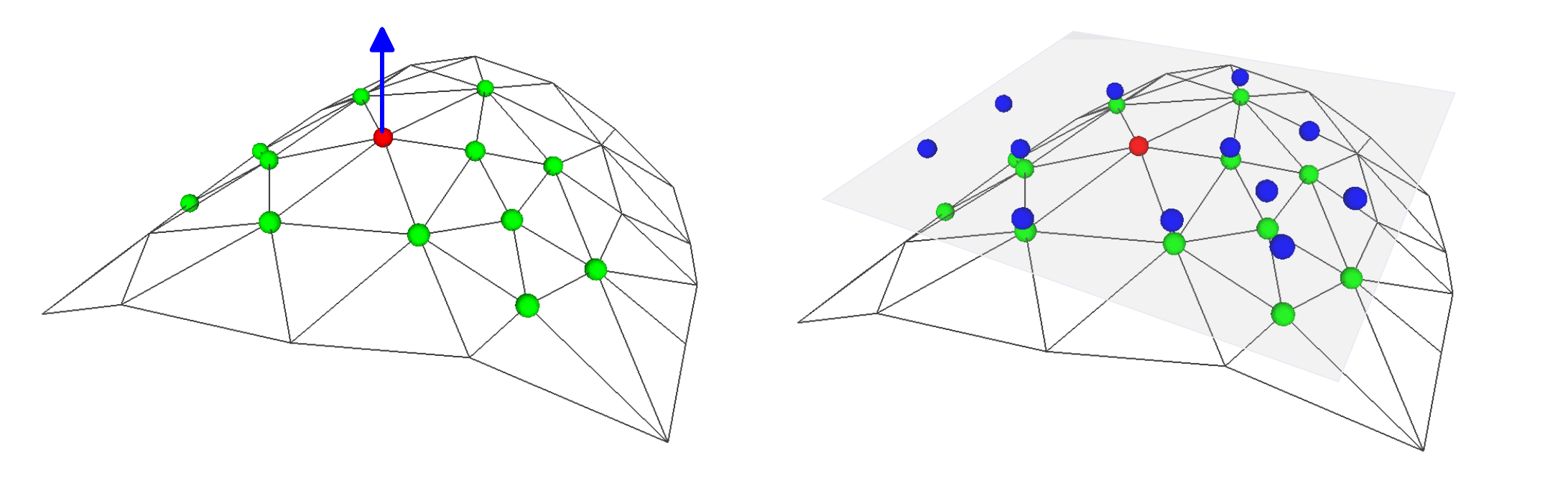} 
\caption{Tangent space mapping for a local region of an input point cloud.}
\label{切平面映射}
\end{figure}

\subsection{Weighted Tangent Smoothing}

{Benefiting} from the initial Poisson resampling, the input point cloud is optimized with uniform distribution and {a} relatively accurate number of points. To ensure the number of points is strictly equal to the specified one and to further optimize the distribution into an isotropic result, we propose a weighted tangent smoothing step that is inspired by the isotropic remeshing~\cite{lv2022adaptively}. In short, it calculates the Voronoi cells for all points and utilizes cells' areas to {weigh} the central displacement. To balance the preservation of sharp features, we {introduce} an adaptive edge resampling option to {meet} the requirements of corresponding applications. 

\textbf{Voronoi Central Displacement.} According to the Centroidal Voronoi Tessellation (CVT)~\cite{du1999centroidal}, the isotropic property can be obtained by Voronoi cell optimization. An implementation has been provided in~\cite{Chen2018La}. However, the drawback lies in its poor computational efficiency. Usually, it takes more than 20 iterations to achieve satisfactory isotropic property~\cite{lv2022intrinsic}. To utilize the advantage of the CVT strategy while improving efficiency, we design a Voronoi central displacement {method} that can be regarded as a weighted point adjustment on the local tangent space.

\begin{figure}[t]
\centering
\includegraphics[width=0.8\columnwidth]{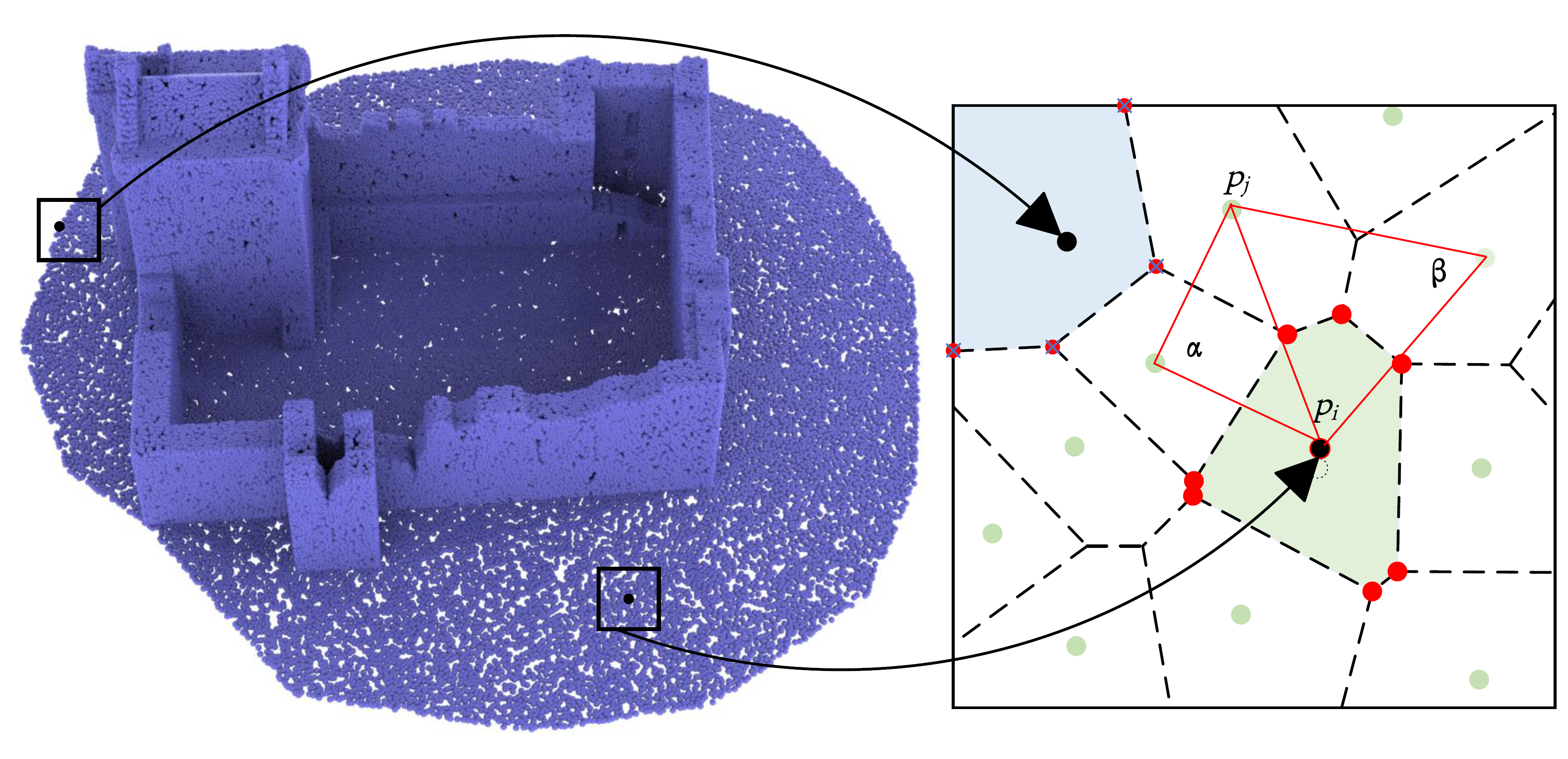}
\caption{As a closed Voronoi cell (green), the related point position is optimized based on the central displacement. For the boundary point with {an} unclosed Voronoi cell (blue), its position should not be changed. The red lines and related labels represent the cotangent weight instance between two points.}
\label{Voronoi图展示}
\end{figure}

At first, we delete {excessive number of} points based on a specified number of points. We calculate distances of all points to their nearest neighbors, {and} then sequentially delete the required number of points starting from the smallest distances. Then{,} we propose the implementation of Voronoi central displacement. Let $p$ represent a point of $P$, we map {the point} and its $k$ neighbors onto the local tangent space, as shown in Figure~\ref{切平面映射}. Since such points are mapped on{to} the 2D plane, computing related 2D Voronoi cell and updating centers becomes extremely convenient. We introduce the cotangent weight to guide the central displacement, represented as

\begin{equation}
\label{e7}
w_{ij}=(\cot\alpha+\cot\beta)/2, w_i = \sum\limits_{p_j\in N_i}{w_{ij}},
\end{equation}
\begin{equation}
\label{e8}
p_i' = \sum\limits_{p_j\in N_i}{\frac{w_{ij}}{w_{i}}p_j},
\end{equation}
where $\alpha$ and $\beta$ are angles between adjacent points (Figure~\ref{Voronoi图展示}), $N_i$ is the adjacent point set of point $p_i$, $p_i'$ is the new position of $p_i$ after the weighted central displacement. We iteratively update point positions by Eq.~\eqref{e7} and Eq.~\eqref{e8}, and the distribution can be optimized with {the} isotropic property.

{Since} the displacement is implemented on the tangent space, some points may not have closed Voronoi cells (blue cell labeled in Figure~\ref{Voronoi图展示}), which could cause their positions to deviate from reasonable ranges. Therefore, we detect these points {that} correspond to unclosed Voronoi cells and keep their positions unchanged to prevent them from escaping {from} the tangent space. It is noteworthy that the above operation relies on a relatively uniform point cloud. Otherwise, during cotangent weight calculation, the convergence speed may decrease, and the weights may become abnormal (e.g., negative values). {Benefiting} from the initial Poisson resampling, the prerequisite of the displacement can be well met to form the basic pipeline of WPD, as shown in Algorithm~\ref{A1_Plane}.

\begin{figure}[t]
\centering
\includegraphics[width=0.95\columnwidth]{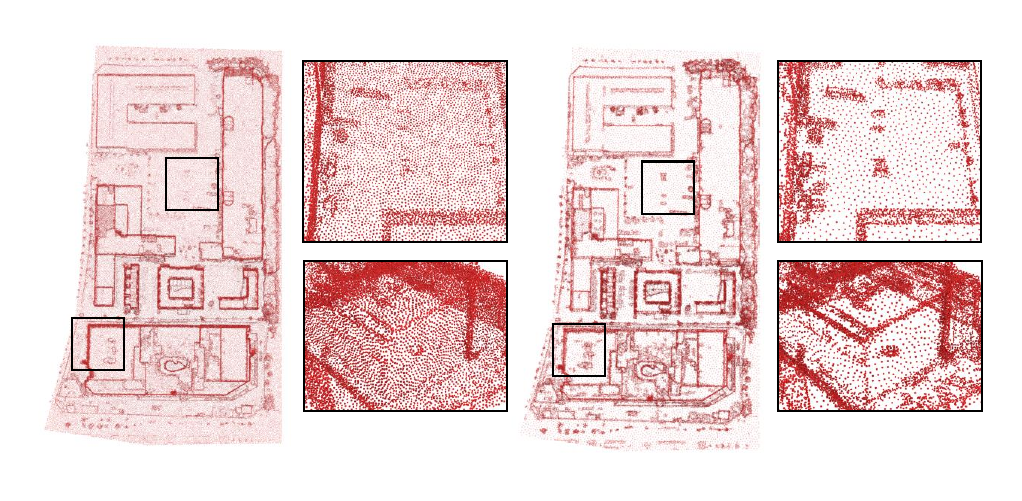}
\caption{Sharp feature keeping in large-scale point cloud. Left: uniform resampling result; Right: uniform resampling result with sharp feature enhancement.}
\label{尖锐特征保持结果}
\end{figure}

\begin{algorithm}[t]
  \caption{The pipeline of WPD}
  \label{A1_Plane}
  \begin{algorithmic}[1]
    \STATE \textbf{Input:} Raw point cloud $P$
    \STATE \textbf{Output:} Resampled point cloud $P'$
    \STATE Implement voxelization for $P$
    \STATE Compute Poisson-disk radius $r$ by Eq.~\eqref{e4}
    \STATE Set $r_e = r/2$ for edge points if used
    \WHILE {$|P|/n > 1.05 \text{ and } |P| < n$}
      \STATE Update $r$ by Eq.~\eqref{e6}
      \STATE Poisson-disk resampling for $P$
    \ENDWHILE
    \FOR {each $p_i \in P$}
      \STATE Extract $k$ neighbors of $p_i$
      \IF {$p_i$ is an edge point and its neighbors have normal points}
        \STATE continue
      \ENDIF
      \STATE Map $k$ neighbors and $p_i$ onto local tangent plane
      \STATE Compute Voronoi cells for all points
      \STATE Update $p_i$ by Eq.~\eqref{e8}
    \ENDFOR
    \STATE Output resampled $P$
  \end{algorithmic}
\end{algorithm}

\begin{figure*}[t]
\centering
\includegraphics[width=\linewidth]{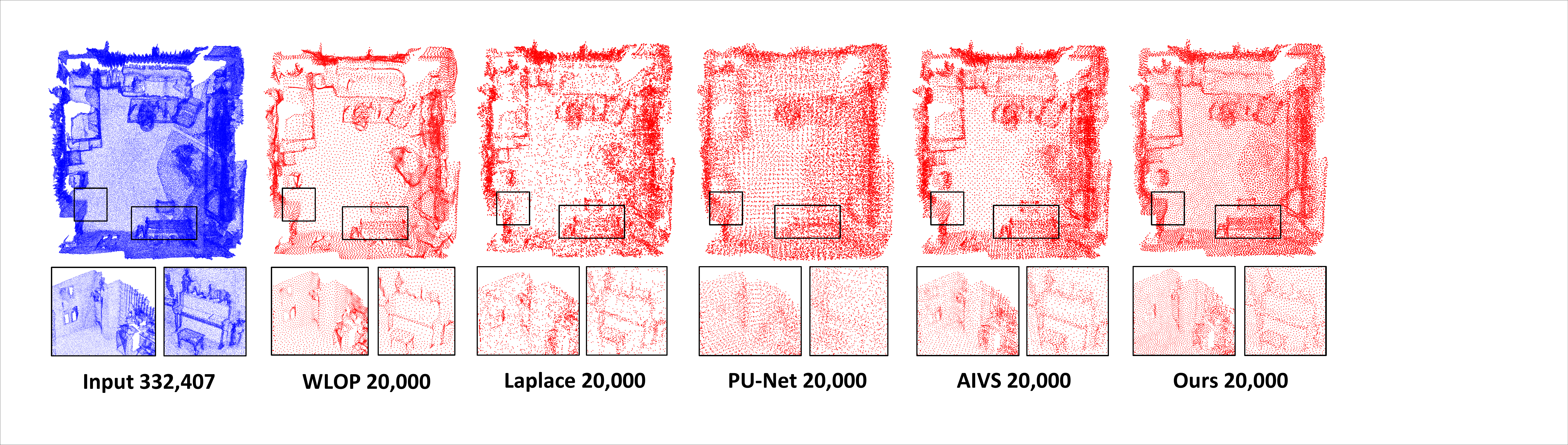} % Reduce the figure size so that it is slightly narrower than the column.
\caption{Comparison with different resampling methods on indoor models of RGB-D dataset. The specified resampling point number is 20,000.}
\label{点云简化结果展示}
\end{figure*}

\begin{figure*}[t]
\centering
\includegraphics[width=\linewidth]{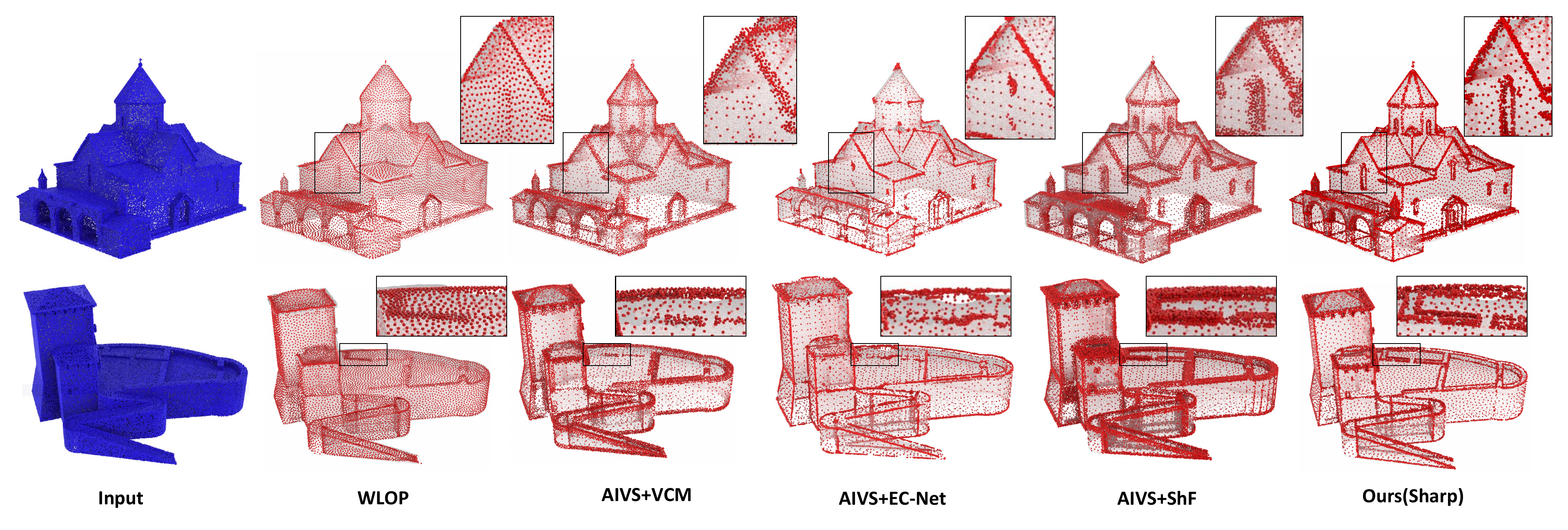} % Reduce the figure size so that it is slightly narrower than the column.
\caption{Comparison of different resampling methods for sharp feature keeping based on architectural models of BuildingNet.}
\label{边缘保持结果展示}
\end{figure*}

\textbf{Sharp Feature Preservation.} For downstream tasks of the resampling, preserving or enhancing sharp features of the point cloud is of significant value for obtaining structural information and conducting high-precision semantic analysis. We provide an optional sharp {feature-keeping} module for the resampling. {Firstly,} we utilize the SOTA solution MSL-Net~\cite{jiao2023msl} to detect edge points. It can handle point clouds with low-level noise, thereby enhancing %which enhances 
its practicality. Then, the point cloud is categorized into normal points and edge points. According to the WPD structure, we set different {radii} for the two kinds of points. To enhance the edge points, we halve their corresponding radius. In this way, the adaptive densities are generated during the initial Poisson resampling. By {the} step of Voronoi central displacement, we assess all edge points and keep their positions if their neighborhood includes normal points. In Figure~\ref{尖锐特征保持结果}, we visualize the sharp feature keeping for large-scale point cloud. Although limited by noise, the enhanced sharp features still reveal the structural boundaries in the scene, {especially} for small-scale semantic objects, which better {maintain} their semantic information. In experiments, we comprehensively evaluate the performance of WPD.

\section{Experiments}

We evaluate the performance of the proposed resampling method. All experiments are processed on a computer equipped with an AMD Ryzen 7 5800H, 16GB RAM, RTX3060, and with {Windows} 11 as its running system and Visual Studio as the development platform. The experiments include the following parts: {(1)} we introduce the selected datasets for the measurement {and comparison}; {(2)} we evaluate the geometric quality for different resampling methods; {(3)} we show some downstream applications based on the resampling method; {(4)} we illustrate the computational efficiency and some other details of our methods for different levels of resampling tasks. The project is released\footnote{github.com/vvvwo/Weighted-Poisson-disk-Resampling}.

\begin{table*}[t]
\centering

\setlength{\tabcolsep}{1.4mm} % 调整列间距以适应内容
\small
\caption{Geometric consistency analysis of different resampling methods on {an} urban dataset. All reported Hausdorff (Haus) and mean distances are divided by 1e+5 for normalization. Bold labels indicate the best results.}
\begin{tabular}{l|cccccccccccccc}
\toprule 
 \textbf{Methods}        & \multicolumn{2}{c}{\textbf{Clustering}}   & \multicolumn{2}{c}{\textbf{WLOP}}& \multicolumn{2}{c}{\textbf{Poission}} & \multicolumn{2}{c}{\textbf{Laplacian}} & \multicolumn{2}{c}{\textbf{PU-Net}} & \multicolumn{2}{c}{\textbf{AIVS}} &\multicolumn{2}{c}{\textbf{Ours}}\\  
 \textbf{Model}  &Haus$\downarrow$ &Mean$\downarrow$  &Haus$\downarrow$ &Mean$\downarrow$ &Haus$\downarrow$ &Mean$\downarrow$ &Haus$\downarrow$ &Mean$\downarrow$ &Haus$\downarrow$ &Mean$\downarrow$ &Haus$\downarrow$ &Mean$\downarrow$ &Haus$\downarrow$ &Mean$\downarrow$\\ 
 \midrule
\textbf{scene645}        & 1930.4     & 43.093   & 772.12  & 44.850 & 421.62   & 44.353  & 1555.9   & 59.497  & 3744.1   & 255.68   & 515.78    & 70.523 & \textbf{421.61}    & \textbf{42.757}\\
\textbf{scene646}         & 2212.6     & \textbf{42.566} & 1465.5  & 54.077 & \textbf{313.19}   & 42.748   & 1777.7   & 69.543  & 4502.8   & 183.93  & 681.54    & 79.500 & 390.36    & 44.512  \\      
\textbf{scene653}          & 992.23     & \textbf{39.394} & 1692.7  & 48.695 & \textbf{317.60}   & 41.903  & 1743.5   & 67.682   & 3987.2   & 186.36  & 557.56    & 83.955 & 402.47    & 41.818 \\
\textbf{scene667}        & 3291.5     & 43.773    & 776.10  & 52.826 & 376.51   & 43.083  & 1860.6   & 69.390  & 3858.4   & 251.52     & 672.25    & \textbf{41.464} & \textbf{327.42}    & 43.415 \\ 
\textbf{scene673}       & 7764.0     & 54.146  & 1155.3  & \textbf{48.889}  & 440.06   & 52.059   & 2168.1   & 71.644   & 3576.4   & 258.24      & 787.73    & 75.335  & \textbf{432.34}   & 51.803\\
\bottomrule          
\end{tabular}

\label{几何一致性对比城市数据}
\end{table*}

\begin{table*}[t]
\centering

\setlength{\tabcolsep}{0.7mm} % 调整列间距以适应内容
\small
\caption{Local and Voronoi density analysis of different resampling methods. %, 
All reported values are divided by 1e+5 for normalization. Bold labels indicate the best results.}
\begin{tabular}{l|cccccccccccccc}
\toprule 
 \textbf{Methods}     &\multicolumn{2}{c}{\textbf{Clustering}} & \multicolumn{2}{c}{\textbf{WLOP}}  & \multicolumn{2}{c}{\textbf{Poission}}  & \multicolumn{2}{c}{\textbf{Laplacian}} & \multicolumn{2}{c}{\textbf{PU-Net}} & \multicolumn{2}{c}{\textbf{AIVS}}   &\multicolumn{2}{c}{\textbf{Ours}}\\ 
 
 \textbf{Model}  &Local$\downarrow$ & Voronoi$\downarrow$  &Local$\downarrow$ &Voronoi$\downarrow$ &Local$\downarrow$ &Voronoi$\downarrow$ &Local$\downarrow$ &Voronoi$\downarrow$ &Local$\downarrow$ &Voronoi$\downarrow$ &Local$\downarrow$ &Voronoi$\downarrow$ &Local$\downarrow$ &Voronoi$\downarrow$\\
 
 \midrule
\textbf{scene645}    &43.09  &33.96  &44.85  &18.47  &44.35  &37.14  &69.54  &71.67  &255.6  &78.76  &70.52  &40.17   &\textbf{42.75}  &\textbf{5.015}\\
\textbf{scene646}    &\textbf{42.56}  &27.42  &54.07  &15.96  &42.74  &33.76  &67.68  &51.23  &183.9  &55.10  &79.50  &28.46   &44.51  &\textbf{6.861}\\   
\textbf{scene653}    &\textbf{39.39}  &25.41  &48.69  &15.42  &41.90  &33.73  &69.39  &42.41  &186.3  &51.76  &83.95  &27.25   &41.81  &\textbf{9.735}\\
\textbf{scene667}    &43.77  &21.11  &52.52  &13.41  &43.08  &24.13  &71.64  &40.10  &251.5  &43.11  &\textbf{41.46}  &25.75   &43.41  &\textbf{6.873}\\ 
\textbf{scene673}    &54.14  &36.73  &\textbf{48.88}  &26.11  &52.05  &39.75  &70.17  &60.07  &258.2  &81.88  &75.33  &42.12   &51.80  &\textbf{10.14}\\
\bottomrule          
\end{tabular}

\label{密度对比室内数据}
\end{table*}

\subsection{Datasets}

The target of our scheme is to implement efficient resampling while considering geometric feature {preservation} on large-scale point clouds. The data we typically handle are scene data, rather than traditional simple models. Based on the target, we collect indoor and outdoor scenes from relevant datasets, including RGB-D dataset \cite{lai2013rgb}, BuildingNet \cite{selvaraju2021buildingnet}, SensatUrban~\cite{hu2021towards}, S3DIS \cite{armeni20163d}, UrbanScene3D~\cite{lin2022capturing}, and UrbanBIS~\cite{Yang2023UrbanBIS}. The RGB-D dataset contains a collection of indoor scenes, including point clouds along with their corresponding RGB values. The number of points in the dataset ranges from 10K to 400K. To verify the effectiveness of our method on large-scale models, we select models with more than 300K points in the experiment. The BulidingNet includes various types of architectural point clouds such as hotels, castles, museums, etc. The number of points in each of these is precisely controlled at 100K. The SensatUrban is an urban-scale point cloud dataset, which contains some large areas from three UK cities. The S3DIS is a classical large-scale indoor dataset for the semantic segmentation study, which includes 13 types of labeled segments such as walls, windows, chairs, etc. We evaluate the performance of some downstream applications on S3DIS models. In {the} following parts, we show the detailed experimental results.

\begin{figure}[t]
\centering
\includegraphics[width=0.83\columnwidth]{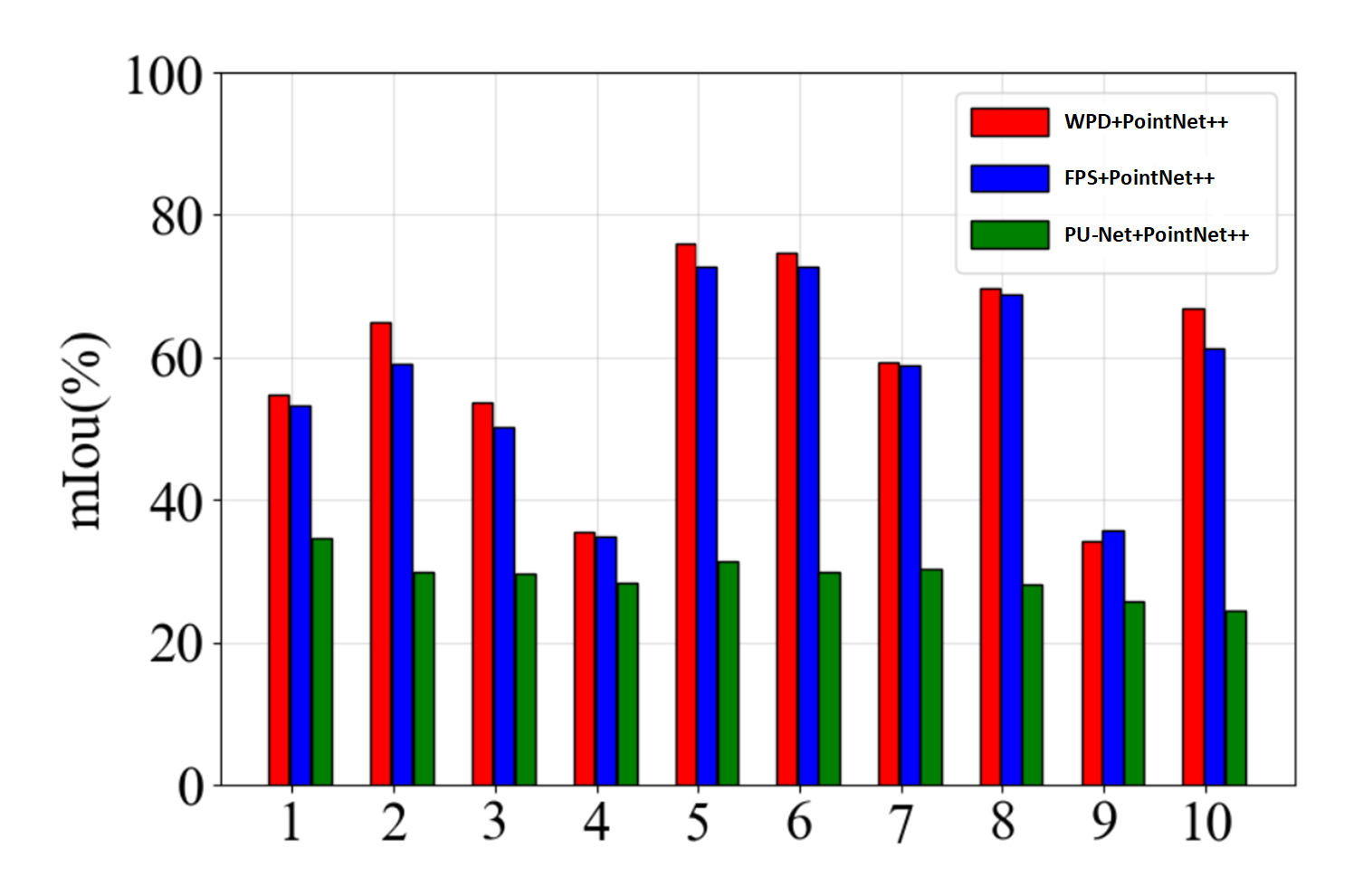}
\caption{Comparison of mIoU values with different resampling{s} based on 10 random{ly} selected S3DIS models.}
\label{分割柱状图}
\end{figure}

\subsection{Geometric Quality}%几何一致性  距离、密度

For geometric quality measurement of the resampling, geometric consistency serves as an important indicator. The reason is that the resampling may change point positions. Such potential impact could disrupt geometric consistency, resulting in {a} resampled point cloud that cannot accurately represent the original one. We conduct a comparison of different resampling methods using {an} RGB-D dataset. The selected metrics include Hausdorff and mean distances, which represent the quality of the geometric consistency. We employ different {comparison methods,} including Clustering~\cite{low1997model}, WLOP~\cite{Huang2009}, Poisson~\cite{corsini2012efficient}, Laplacian~\cite{luo2018uniformization}, PU-Net~\cite{yu2018pu}, and AIVS~\cite{lv2021approximate}. In Table~\ref{几何一致性对比城市数据}, we report the results. Although clustering and Poisson methods can achieve better results on some point clouds, neither of them can precisely control the resampling point number. In contrast, our method achieves relatively stable quality of geometric consistency in test samples.

\begin{table}
\centering
\setlength{\tabcolsep}{0.8mm} % 调整列间距以适应内容
\small
\caption{Time cost report for resampling methods in large-scale point clouds. Output point number is set to 10k.}
\begin{tabular}{l|ccccc}
\toprule 
 \textbf{Input$\backslash$Methods}       & \textbf{WLOP}   &\textbf{Laplacian} & \textbf{PU-Net} & \textbf{AIVS}  &\textbf{Ours}  \\ 
 \midrule
\textbf{2,000K-200K}       & $>$10min    & $>$10min  & $>$1min   & $>$10min & \textbf{12.7s}\\
\textbf{200K-50K}       & $>$1min    &$>$1min  & 34s   & $>$1min & \textbf{3.7s}\\
\textbf{$<$50K}             & \textbf{0.25s}   & 23s     & 2.1s  & 12s  & 0.42s \\
\bottomrule          
\end{tabular}
\label{城市数据网格重建结果对比1}
\end{table}

\begin{figure*}[t]
\centering
\includegraphics[width=0.98\linewidth]{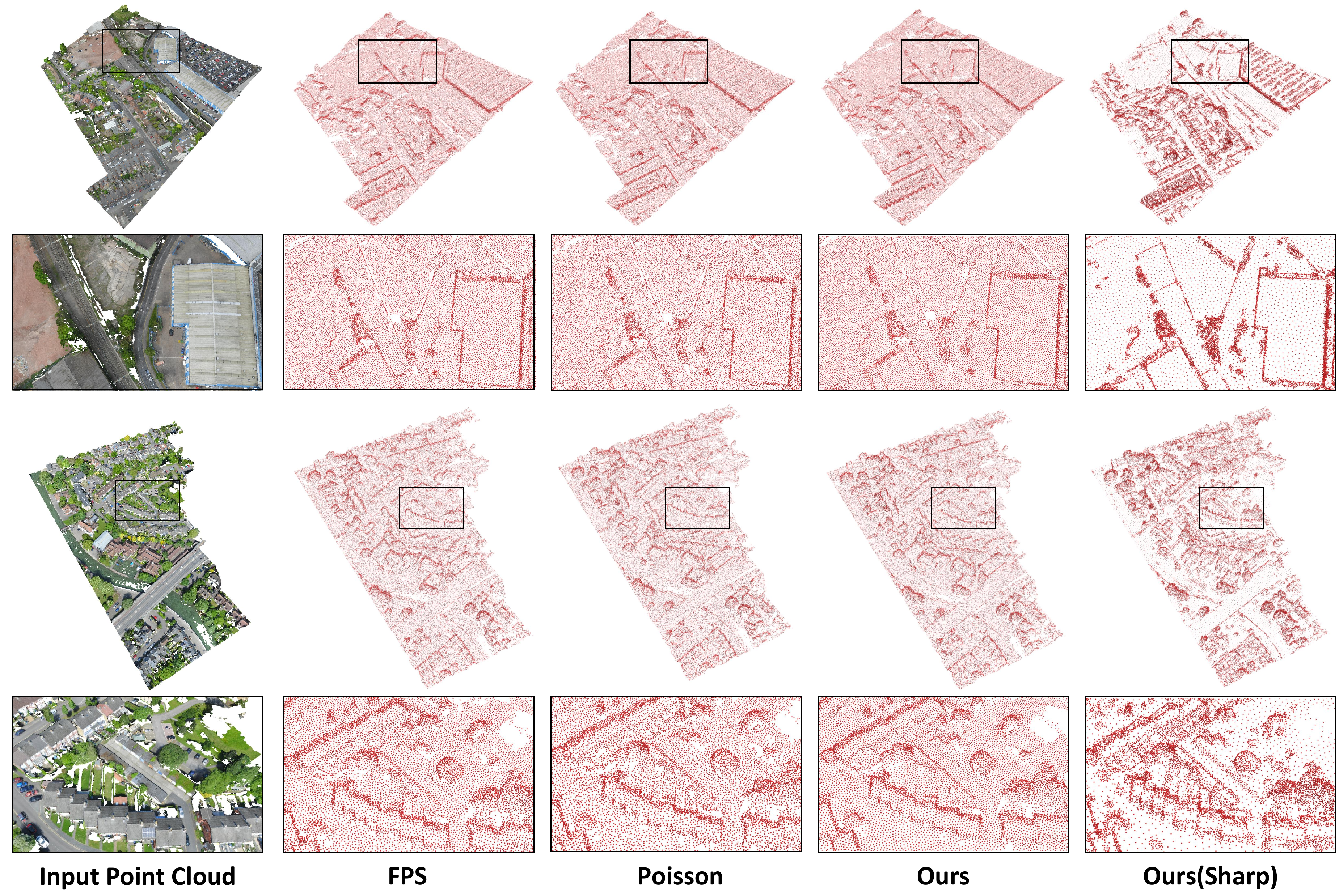}
\caption{Comparison of different resampling methods for urban scenes of SensatUrban. The specified resampling point number is 100K. Our resampling method (sharp) retains more geometric details.}
\label{UrbanResampling}
\end{figure*}

Another important indicator of geometric evaluation is the point-based uniformity or isotropic property. {It is worth noting that} defining the neighborhood structure of point clouds is a fundamental issue in point cloud analysis. Uniform distribution in the neighborhood allows the point cloud to cover larger continuous areas with fewer points while facilitating the definition of accurate adjacent regions. To evaluate the uniformity of the point cloud, we employ two kinds of measurement: local and Voronoi density errors. The local density is based on the analysis of neighbor point densities, which are represented as
\begin{equation}
\begin{array}{c}
    {d_{local}}=\max \{\overline{{D}}\}-\min \{\overline{{D}}\}, \overline{d_i}\in \overline{{D}}, \\ \\
    \overline{{{d}_{i}}}=\sum\limits_{{{p}_{j}}\in K({{p}_{i}})}{d({{p}_{i}},{{p}_{j}})}/k,
\end{array} 
\label{局部密度dloc计算公式}
\end{equation}
where $\overline{d_i}$ represents the average distance between the point $p_i$ and its $k$-nearest neighbor set $K(p_i)$ ($k = 6$ to fit isotropic property), $\overline{D}$ is the set of average distances based on the point cloud. The difference between the maximum and minimum values of $\overline{D}$ roughly reflects the uniformity of point distributions. For accurate neighborhood analysis, we compute the Voronoi region {of} the $k$-nearest neighbor set based on Eq.~\eqref{局部密度dloc计算公式}. In Table~\ref{密度对比室内数据}, we report the results of uniformity measurement for different resampling methods. Our method significantly achieves better point-based distribution.

Preserving or enhancing sharp features during resampling is an additional metric for geometric quality assessment. Sharp features carry important structural information, which have significant value for surface estimation and semantic learning. %For sharp feature keeping, we report a specific experimental test based on BuildingNet. 
To evaluate sharp feature preservation, we report the results of a specific experimental evaluation on BuildingNet. Some methods with sharp feature keeping are used for comparison, as shown in Figure \ref{边缘保持结果展示}. For the WLOP scheme, we set the neighborhood scale size to 0.03 based on the BuildingNet model, which achieves the best performance for sharp feature keeping in experience. For the AIVS scheme, it employs the Voronoi-based feature estimation (VCM)~\cite{merigot2010voronoi} to balance the sharp feature keeping and isotropic resampling. We set different resampling rates for sharp feature points (0.88) and normal points (0.12) and output {the} same point number in {the} final results, which are used to balance the feature keeping and uniformity by experience. With {the} same resampling point number (17,000), our method keeps more sharp points for the building's outlines, doors, and windows, while retaining fewer but evenly distributed points for flat areas such as walls and ground. To some extent, our method achieves sharp feature enhancement {with the} dynamic resampling strategy. In Figure~\ref{UrbanResampling}, we show more resampling results based on urban models of SensatUrban, UrbanScene3D, and UrbanBIS. Most existing resampling methods cannot achieve results with acceptable time cost ($>1$ minute in Table \ref{城市数据网格重建结果对比1}). As the most efficient method currently, Poisson-disk resampling implemented by MeshLab takes more than 30\% average error rate in controlling sampling points. In contrast, our method has significant advantages in controlling the number of points and preserving sharp features.

\subsection{Applications}

To further demonstrate the performance of the proposed resampling method, we present two related downstream applications: mesh reconstruction and semantic analysis.

\begin{figure*}[t]
\centering
\includegraphics[width=0.98\linewidth]{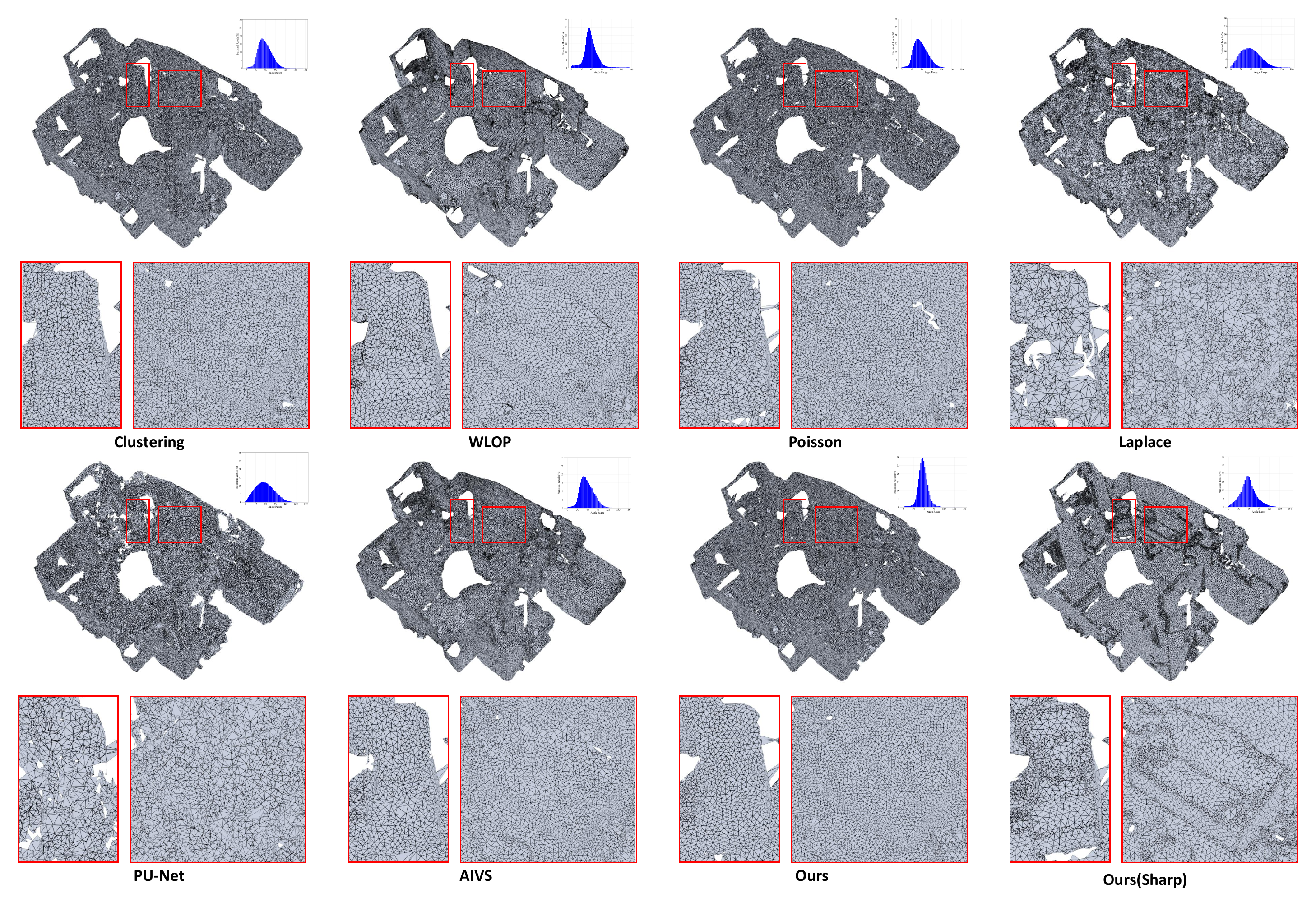}
\caption{Comparison of reconstructed meshes with statistical histograms of inner angles using different resampling methods based on S3DIS models. WPD can achieve better isotropic property.}
\label{网格重建结果展示}
\end{figure*}

\textbf{Mesh Reconstruction.} Constructing 3D surfaces from point clouds is a fundamental challenge in computer graphics and 3D vision. In explicit surface reconstruction, particularly when employing Delaunay triangulation, the uniform distribution of point clouds plays a critical role in mesh generation. An isotropic distribution can enhance the precision of neighborhood definition. Based on the S3DIS scenes, we evaluate the quality of triangulation with different resampling methods. For a fair comparison, we set the resampling point number to 50,000 for all models. The triangulation is implemented by Geomagic, {which is} a commercial mesh generation software. A comparative example is shown in Figure \ref{网格重建结果展示}. Intuitively, the reconstructed mesh exhibits better isotropic property by our resampling scheme.

\textbf{Semantic Analysis.} For complex point cloud scenes, uniform distributions of points are helpful for semantic analysis. The reason is that the uniform distribution helps in learning local geometric and semantic features. The adjacency relationship between points is more easily determined. To verify the impact of resampling on semantic analysis, we conducted corresponding experiments. We select PointNet++~\cite{qi2017pointnet++} to be the backbone and resample point clouds with different methods{, including} FPS by default~\cite{qi2017pointnet++}, PU-Net~\cite{yu2018pu}, and WPD. Based on the indoor models of {the} S3DIS dataset, we compare their segmentation accuracy. For the quantification of the performance improvement of our method, we computed the mIoU metrics for 10 scenes from S3DIS, as shown in Figure~\ref{分割柱状图}. The proposed WPD resampling improves the accuracy of semantic analysis.

\subsection{Discussion}

Compared to the original Poisson-disk sampling method, our initial Poisson resampling approach enhances the control accuracy of sampling points without incurring additional computational overhead. By leveraging weighted tangent smoothing, our resampling method yields isotropic results from large-scale point clouds. As demonstrated by its efficiency and precision across various resampling tasks of differing scales, our method stands capable of accomplishing them with remarkable accuracy, as shown in Table \ref{城市数据网格重建结果对比1}. Original Poisson-disk resampling cannot precisely control the point number, which is not reported. Even when the input point count reaches the order of one million, our method's time cost remains manageable, typically around 10 seconds. Conversely, under identical conditions, both WLOP and AIVS incur time overheads exceeding 10 minutes.  

As mentioned in downstream applications, the resampling is useful. For mesh reconstruction, our resampling method can optimize local neighborhoods of all points, which {improves} the accuracy of topological information. For semantic analysis, the resampled point cloud takes uniform density, {which} is helpful for semantic feature learning. Such improvements {benefit} from the isotropic property. The above experiments demonstrate that our method possesses performance comparable to replacing Farthest Point Sampling (FPS) and Poisson-disk resampling comprehensively. It can provide accurate point number control and density optimization for large-scale point clouds, as shown in Figure~\ref{UrbanResampling}.

\section{Conclusions}

In this paper, we propose a weighted Poisson-disk resampling method for point number control and isotropic optimization on large-scale point clouds. Firstly, we employ a voxel-based analysis to establish an initial Poisson resampling, which improves the accuracy of Poisson-disk radius estimation. Then, a weighted tangent smoothing step is used to optimize {the} isotropic property while optionally keeping sharp features. The proposed resampling method balances efficiency and accuracy, {which is} particularly advantageous for processing large-scale point cloud data. Experiments show that our method can handle various resampling tasks and improve their performance.

\bibliography{aaai25}

\end{document}